\documentclass[12pt]{mmcp_abstract}

\usepackage{graphicx}
\usepackage{amsmath}
\usepackage{amsfonts}
\usepackage{amssymb}
\usepackage{subfig}
\title{Accelerating Domain-aware Deep Learning Models with Distributed Training}
\author{Aishwarya Sarkar, Chaoqun Lu and Ali Jannesari}

\address{Iowa State University, USA \\
\{asarkar1, clu, jannesar\}@iastate.edu
}

\begin{document}
\begin{sloppypar}
\vskip0.5cm
\section{Abstract and Motivation}
Recent advances in data-generating techniques led to an explosive growth of geo-spatiotemporal data. In domains such as hydrology, ecology, and transportation, interpreting the complex underlying patterns of spatiotemporal interactions with the help of deep learning techniques hence becomes the need of the hour. However, applying deep learning techniques without domain-specific knowledge tends to provide sub-optimal prediction performance. Secondly, training such models on large-scale data requires extensive computational resources. To eliminate these challenges, we present a novel distributed domain-aware spatiotemporal network that utilizes domain-specific knowledge with improved model performance. Our network consists of a pixel-contribution block, a  distributed multiheaded multichannel convolutional (CNN) spatial block, and a recurrent temporal block. We choose flood prediction in hydrology as a use case to test our proposed method. From our analysis, the network effectively predicts high peaks in discharge measurements at watershed outlets with up to 4.1x speedup and increased prediction performance of up to 93\%. Our approach achieved a 12.6x overall speedup and increased the mean prediction performance by 16\%. We perform extensive experiments on a dataset of 23 watersheds in a northern state of the U.S. and present our findings.

Domain-aware deep learning (DL) models do not solely rely on data in hand. Having some insight about the targetted domain is what makes them stand out. Without explicit constraints, DL models are prone to make erroneous predictions. Research shows that these models outperform their data-driven counterparts in various applications such as human pose estimation \cite{8064661}, studying animal models \cite{zhong2021dika}, breast cancer diagnosis \cite{chen2021domain}, and seismic inversion \cite{sun2021physics} among others. To make a DL model domain-aware, researchers have tried to closely integrate them with their domain-specific counterparts \cite{sarkar2020hydrodeep}. Despite these advances, the bottleneck of efficiently training, testing, and deploying these models still remains a challenge. The recent breakthrough in high-performance computing systems shows immense potential to make domain-aware DL models overcome bottlenecks by making them scalable, and cost-effective \cite{yu2022towards, yu2022spatl}. 

Global floods and extreme rainfall events have surged by more than 50\% this decade and are now occurring at a rate four times higher than in 1980. Increasing evidence shows the huge potential of predicting and mitigating flood occurrences in reducing yield loss, improving disaster risk management, and benefiting the environment \cite{hess-11-96-2007, https://doi.org/10.1002/hyp.333, WHEATER2009S251}. This work tackles the challenges in DL-based pixellated spatiotemporal flood prediction models by introducing a novel approach to somewhat eliminate its black box-like nature. It exploits domain knowledge that comes from pixelated inputs by learning their pixel-specific spatiotemporal contributions. We further propose a novel distributed training approach to accelerate and optimize the training time of a pixellated domain-aware spatiotemporal network. In summary, our contributions are:
\begin{enumerate}
\item A domain-aware distributed spatiotemporal network (Dom-ST) with a novel pixel-contribution block (Pix-Con) to compute pixel-specific weights that transforms spatiotemporal inputs based on their local contribution to the water discharge (Figure 1).
\item A partitioning module in the Pix-Con block that partitions spatiotemporal pixels dynamically and distributes them to multiple devices (Figure 1).
\item A novel domain-guided distributed training approach to accelerate training (Figure 2).
\item An approach to utilize domain-significant temporal input measurements (the target day) to increase domain awareness in Dom-ST.
\item Competitive results in improving flood prediction by upto 93\% and an overall speedup of 12.6x with domain-guided distributed training,
\end{enumerate}
\begin{figure*}[!t]
    \centering
    \includegraphics[width=\textwidth]{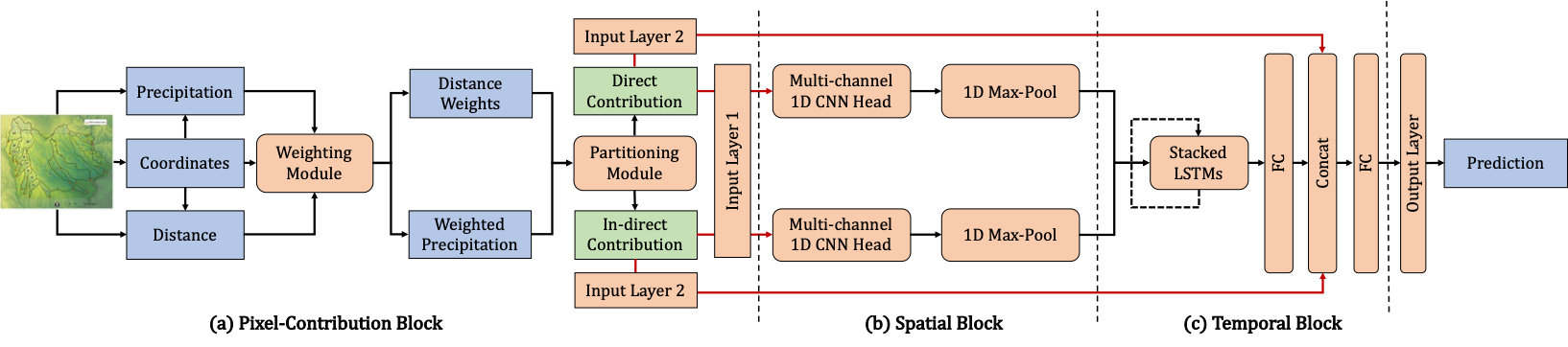}
    \caption{The overall workflow of the proposed distributed spatiotemporal network Dom-ST with (a) pixel-contribution (Pix-Con) block, (b) spatial block with parallel convolution heads, and a recurrent (c) temporal block. The blue boxes represent data, and the orange boxes represent various modules of Dom-ST. The arrows show the data flow during inference.}
    \label{fig:my_label}
\end{figure*}
\section{Distribution Strategy}
Since the climate and other environmental factors of a region, such as soil, land cover, temperature, and human impact, vary from one location to another, one global hydrological model, if trained, will either give suboptimal results or will be computationally extensive. Model performance can be improved further from the knowledge gained by modeling targetted regions. Besides, data scarcity in remote areas is a challenge of its own which can be tackled to some extent by transfer learning approaches \cite{sarkar2021transfer}. Due to the huge amount of complex multi-dimensional spatiotemporal data, training a hydrological model based on deep learning approaches on a single device requires more resources in terms of time and memory. Our distributed training approach tackles this challenge (Figure 2). Firstly, the training data of the targetted site is split into 23 regions or watersheds in the input pipeline (I.P.), which then distributes the set of watersheds to multiple nodes (Figure 2.a.), where each node is responsible for a single watershed. The I.P. also creates model replicas and distributes them for each watershed to corresponding nodes. Secondly, in each watershed node, we partition the model and offload it to multiple GPUs by exploiting the parallel architecture of our spatial block (Figure 2.b.). Each head of the CNN is separated and trained on a single separate GPU making the heads train parallelly, thus optimizing the training time. Once the forward passes of the spatial blocks are completed, the outputs are transferred to the next GPU, which has the temporal block containing the stacked LSTM layers and the following final layers of the model. The temporal GPU also receives the second input of the model directly. 
\begin{figure*}[!t]
    \centering
    \includegraphics[width=\textwidth]{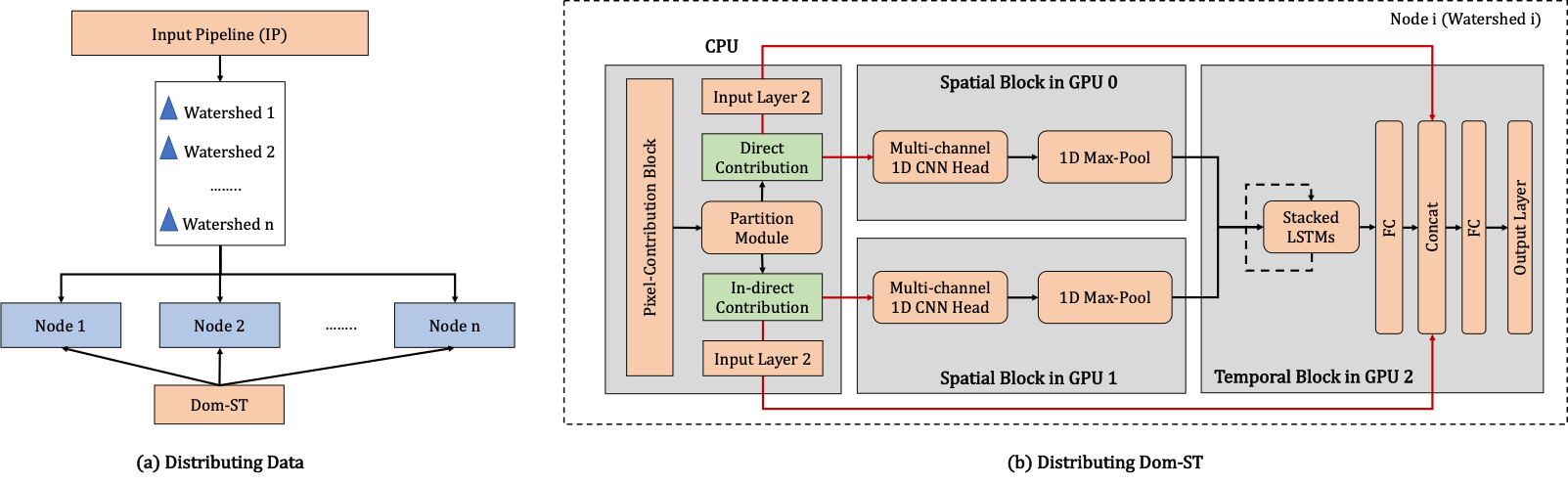}
    \caption{(a) The input pipeline distributes chunks of data (watersheds) and creates Dom-ST replicas on multiple nodes (Left). (b) Each node runs distributed Dom-ST using the host (CPU) and multiple devices (GPU) (Right).}
    \label{fig:my_label}
\end{figure*}

We use pixellated daily precipitation from the Climate Research Unit and distances of each pixel from the nearest water source as inputs and U.S. Geological Survey's daily discharge measurements \cite{jones-et-al:scheme} as labels. We selected Nash-Sutcliffe efficiency (NSE) as our model evaluation metric \cite{moriasi2007model}. Our baseline model consists of only the spatial and temporal blocks. The spatial block consists of singlehead one-dimensional multichannel 1D-CNN without any Pix-Con block, which implies that the baseline is purely data-driven with no domain knowledge. We refer to this as \textit{Singlehead}. The baseline runs on a single GPU. We explore how having an additional input of the target day's precipitation outperforms the baseline's performance. Figure 3 shows that in almost all the watersheds (91\%), feeding the target day's precipitation in the final layers of the model (\textit{Singlehead}$(+P)$) improves the performance. The target day's precipitation, the primary contributing factor of flash floods, provides temporally fine-grained data to Dom-ST as a domain-specific cue to improve the prediction. Next, we evaluate our proposed Dom-ST, which has a Pix-Con block, a distributed multihead spatial block, and a temporal block that receives the target day's inputs. For comparison, we refer to it as \textit{Distributed-Multihead(+P)} in Figure 3. From our results, \textit{Distributed-Multihead(+P)} outperforms both \textit{Singlehead} and \textit{Singlehead}$(+P)$ in almost all the watersheds. This supports that Dom-ST is more domain-aware than its counterparts. To evaluate the input pipeline strategy, we compare the computation time of \textit{Singlehead+}$P$ and \textit{Distributed-Multihead(+P)}, each running sequentially (S) and then parallelly (IP-D). Table 1 shows the total training time of 23 watersheds in each case.
\begin{figure*}[!h]
    \centering
    \includegraphics[width=0.7\textwidth]{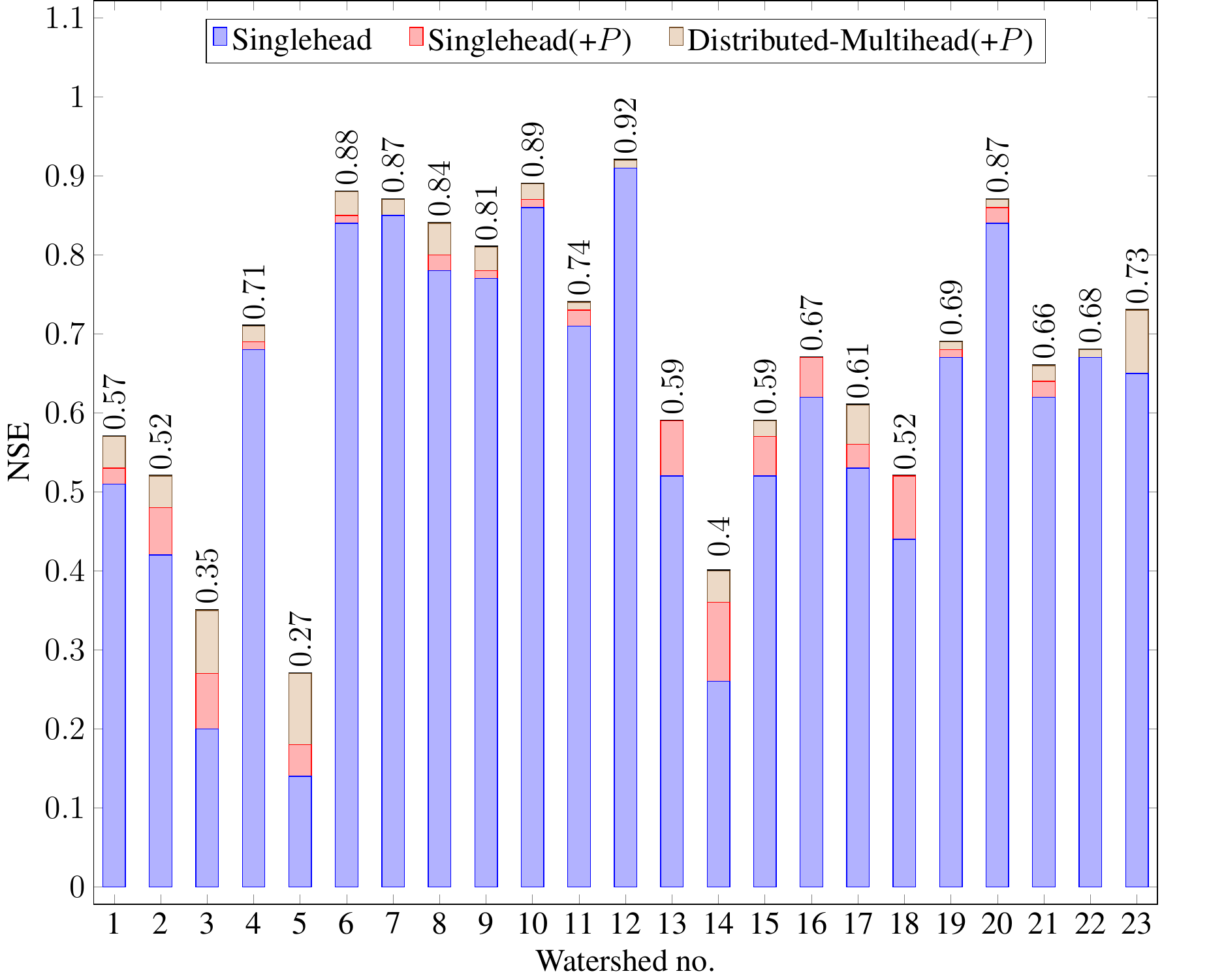}
    \caption{Comparison of prediction performance in terms of NSE \cite{nash1970river} between 3 approaches - singlehead, singlehead with extra precipitation inputs (Singlehead$(+P)$), and distributed-multihead with extra precipitation inputs (Distributed-Multihead$(+P)$).}
    \label{fig:my_label}
\end{figure*}
\begin{table}[h!]
\caption{Evaluation of the input pipeline strategy (IP-D).}
\begin{center}
\scalebox{0.7}{\begin{tabular}{|c|c|c|c|}
\hline
\textbf{Approach} & \textbf{Time (S)} & \textbf{Time (IP-D)} & \textbf{Speedup} \\
\hline
Singlehead($+P$)& 9.96 hours & \textbf{1.18} hours & 8.5x\\
\hline
Distributed-Multihead($+P$) & 5.49 hours & \textbf{0.44} hours & 12.6x\\
\hline
\end{tabular}}
\label{tab1}
\end{center}
\end{table}

\section{Conclusion}
We propose a novel distributed training approach to accelerate a domain-aware spatiotemporal network. Through rigorous experiments and ablation studies, we further explore how domain knowledge drives a model's prediction and can also be utilized in finding an approach for distributed training by exploiting pixellated spatiotemporal data to infuse pixel-level contribution. Our findings indicate that the use of Dom-ST leads to an average speedup of $2.02$x across all the analyzed watersheds. The pixel-contribution block in Dom-ST utilizes domain-knowledge and improves the model's prediction performance. Our experiments achieved the highest individual watershed speedup of up to $4.11$x and an overall speedup of $12.6$x with the highest increase in individual NSE by $93\%$.
\bibliographystyle{unsrt}
\bibliography{main}
\end{sloppypar}
\end{document}